\documentclass{article}
\usepackage{graphicx}
\usepackage{amsmath}
\usepackage{amssymb}
\usepackage{xcolor}
\usepackage{geometry}[margin=1in]
\usepackage{hyperref}
\usepackage{physics}
\usepackage[labelfont=bf]{caption}
\hypersetup{
    colorlinks,
    linkcolor={blue!50!black},
    citecolor={blue!50!black},
    urlcolor={blue!80!black}
}
\usepackage{authblk}
\usepackage[capitalize,noabbrev,nameinlink]{cleveref}

\makeatletter

\renewcommand\AB@affilsepx{, \protect\Affilfont}
\renewcommand\Affilfont{\small}
\makeatother

\title{Exploiting Inductive Biases in Video Modeling through Neural CDEs}
\author[1]{Johnathan Chiu}
\author[1]{Samuel Duffield}
\author[1]{Max Hunter-Gordon}
\author[1]{\\Kaelan Donatella}
\author[1]{Maxwell Aifer}
\author[1,2]{Andi Gu}
\affil[1]{\href{https://normalcomputing.ai/}{Normal Computing}} \affil[2]{Harvard University}
\date{}

\usepackage{titling}
\begin{document}

\maketitle

\begin{abstract}
We introduce a novel approach to video modeling that leverages controlled differential equations (CDEs) to address key challenges in video tasks, notably video interpolation and mask propagation. We apply CDEs at varying resolutions leading to a continuous-time U-Net architecture. Unlike traditional methods, our approach does not require explicit optical flow learning, and instead makes use of the inherent continuous-time features of CDEs to produce a highly expressive video model. We demonstrate competitive performance against state-of-the-art models for video interpolation and mask propagation tasks.
\end{abstract}

\section{Introduction}

A bedrock principle in statistics and learning theory is the notion of \emph{inductive biases} \cite{mitchell1980}. Loosely defined, inductive bias is the set of \textit{a priori} (i.e., not learned from training data) structural assumptions that a model uses to make predictions on unseen data. Identifying the correct inductive biases to encode into models has played a pivotal role in the modern successes of machine learning. A classic example of this is the development of convolutional neural networks (CNNs) for image recognition tasks. The special structure of CNNs implicitly encodes an inductive bias of locality and translational invariance, which has been recognized \cite{bronstein2021geometric} as a critical factor in their initial remarkable success \cite{lecun1998} on image data over more heavy-handed methods such as fully-connected neural networks. Indeed, a cursory review of the history of machine learning research reveals that a significant driving force for progress has been the identification of novel inductive biases for an array of tasks, including, to name just a few, language modeling \cite{vaswani2023attention}, computational chemistry \cite{gilmer2017neural}, physics \cite{sanchezgonzalez2019hamiltonian,cranmer2020lagrangian}, and modeling relational data \cite{battaglia2018relational}.

\begin{figure*}
    \centering
    \includegraphics[width=\textwidth]{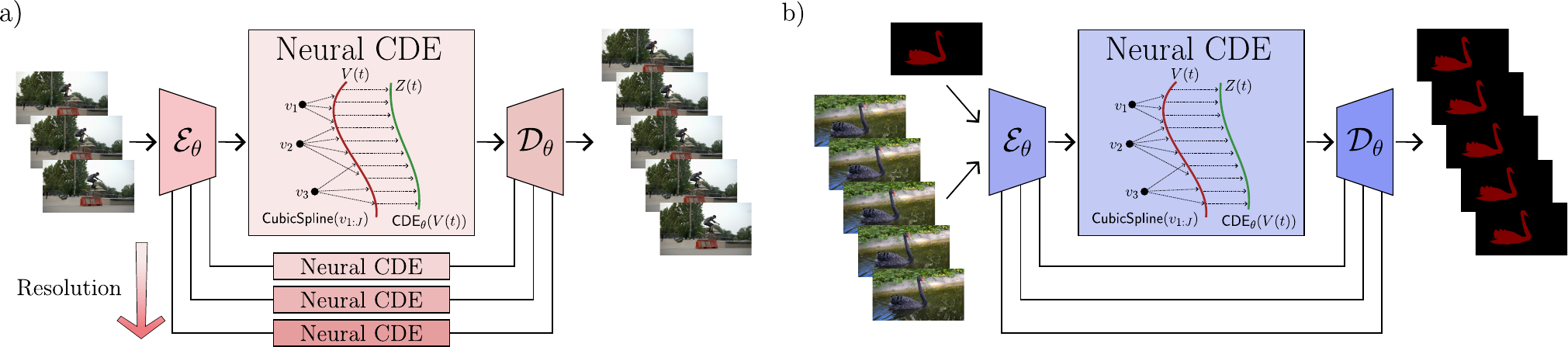}
    \caption{Diagram of the architecture employed in this work for the tasks of (a) frame interpolation and (b) mask propagation. In both cases a sketch of the neural CDE framework is shown. In the case of frame interpolation an incomplete video is fed into an encoder who's output is fed into a neural CDE. The neural CDE propagates the latent vector of the frame after the action of a cubic spline. The output of the neural CDE is then fed into a decoder which maps the output latent space vector to a video with frames which have been interpolated. Multiple neural CDEs are applied at cascading (latent) resolutions, thus the overall architecture can be interpreted as a \textit{continuous-time U-Net}. For the mask propagation architecture we use splines across resolutions with the CDE applied only at the smallest latent space, reminiscent of the skip connections in the original U-Net \cite{ronneberger2015u}.
    }
    \label{fig:scheme}
\end{figure*}

The compatibility of differential equations with machine learning is well-established, including Langevin-based Monte Carlo methods \cite{neal2011mcmc}, diffusion models \cite{sohl2015deep, song2020score} and neural differential equations \cite{chen2018neural, kidger2022neural}. The aforementioned methods most commonly use continuous-time dynamics to aid the modelling of data that is not assumed to be temporal. In this work, we focus on the use of continuous-time dynamics to encode inductive biases in temporal data. Specifically, in the context of video modeling, the temporal continuity present in videos offers a rich source of inductive bias. Recent work on video modeling has exploited this temporal continuity by assuming the form of a latent ordinary differential equation (ODE) parameterized by a neural network \cite{yildiz2019ode2vae, park2021vid, kanaa2021simple}. An inherent issue with using ODEs for temporal data is that the entire ODE trajectory is fully determined given an initial value, this is typically mitigated by applying discrete-time updates of a hidden state \cite{rubanova2019latent, park2021vid}. This important issue points to using a more general framework to model temporal continuity.

In this work, we use neural controlled differential equations (CDEs), which learn an expressive mapping entirely in continuous-time, capturing and leveraging the temporal coherence present in videos. Neural CDEs were first applied to the problem of irregular time-series modeling~\cite{kidger2020neural}, where the trajectory is adjusted based on subsequent observations, and not only the initial condition like in ODE-based methods. 

Reminiscent of the popular and successful U-Net \cite{ronneberger2015u} architecture we apply the neural CDEs at varying resolutions to increase the spatial expressibility of the continuous-time dynamics, depicted in \cref{fig:scheme}. To benchmark our method, we focus on two tasks: frame interpolation, which involves synthesizing intermediate images between two or more input frames, and mask propagation, where the objective is to track the trajectory of a given mask in the original video. We demonstrate the effectiveness of neural CDEs on these two tasks, where we find comparable results to the state-of-the-art. 

\section{Related Work}

As mentioned, neural ODEs have been used in modeling both video \cite{yildiz2019ode2vae, park2021vid, kanaa2021simple} and general time series data ~\cite{chen2018neural, li2020scalable}. Similarly, diffusion models \cite{sohl2015deep} have proved extremely successful at image generation and have even been adapted to videos \cite{ho2022video}. Although video diffusion models may formulate the noising process in continuous-time they typically leave the temporal aspect of the frames in discrete-time combining ideas from sequence models like recurrent neural networks or transformers. The combination of continuous-time latent temporal representations and artificial continuous-time noising from diffusion models represents a promising area for future research.

\paragraph{Continuous-time Video Modeling} 
Refs. \cite{yildiz2019ode2vae, park2021vid, kanaa2021simple} appear to be the work most closely related with what we present here. In \cite{yildiz2019ode2vae}, the authors use two variational auto-encoders to learn a latent space, one for the position and the other for the velocity, with the acceleration of objects in a video defined by a learnable ODE. This information is then used to solve several video prediction tasks. In \cite{kanaa2021simple} the authors lay the foundation for latent space models using ODEs for video tasks. They propose using the learned latent space evolution of videos to predict later frames. Finally, in \cite{park2021vid}, a neural ODE approach is combined with pixel level video processing resulting in an ODE with convolutional gated recurrent units (GRUs), resulting in a model that can predict video frames at any given time steps. One could argue that this model can fall under the CDE definition in which the GRUs used to reconstruct continuous-time dynamics represents the ``control'' mechanism.

As mentioned previously, and unlike the aforementioned video techniques, we learn continuous-time dynamics at varying resolutions taking inspiration from image U-net architectures \cite{ronneberger2015u}. A somewhat related continuous-time U-Net framework was proposed in \cite{rahman2022u} in the very different setting of solution operators for partial differential equations and without the use of CDEs.

\begin{figure*}[!ht]
    \centering
    \includegraphics[width=0.8
\textwidth]{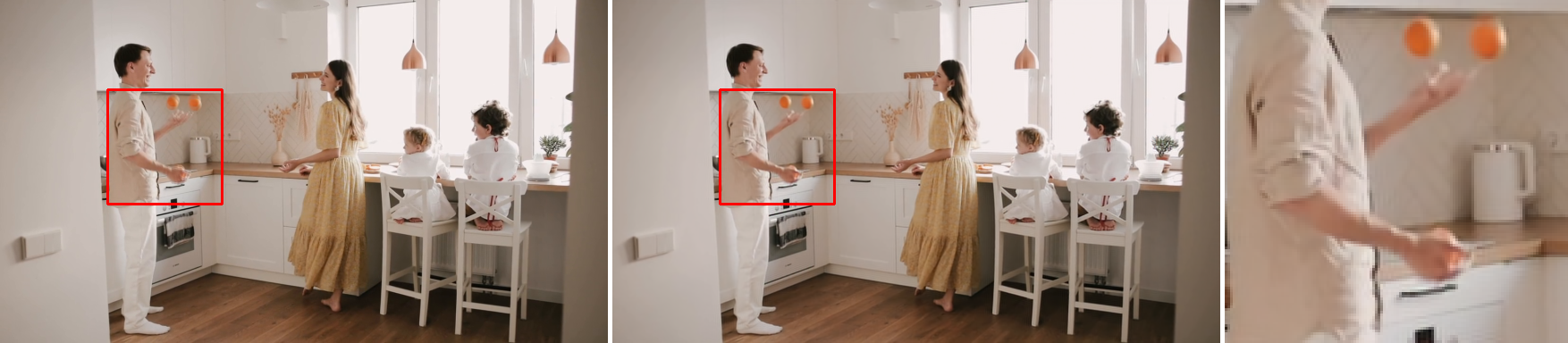}
    \includegraphics[width=0.8\textwidth]{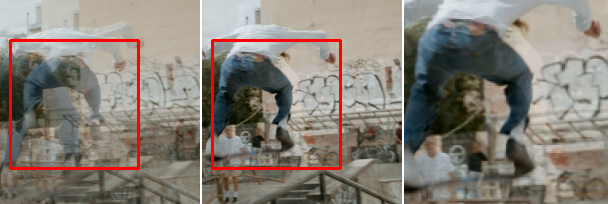}
    \includegraphics[width=0.8\textwidth]{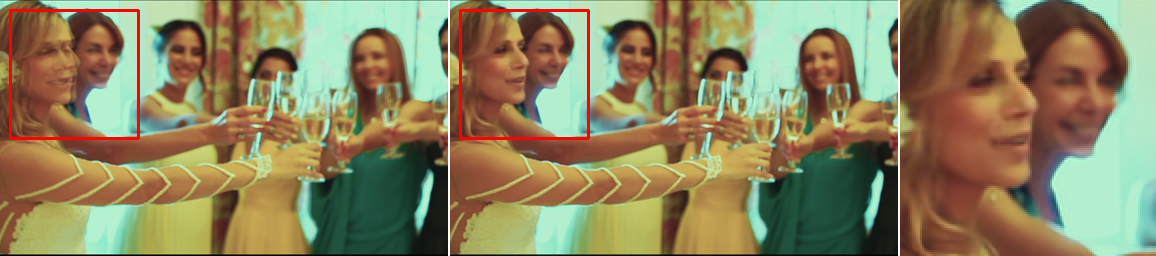}
    \includegraphics[width=0.8\textwidth]{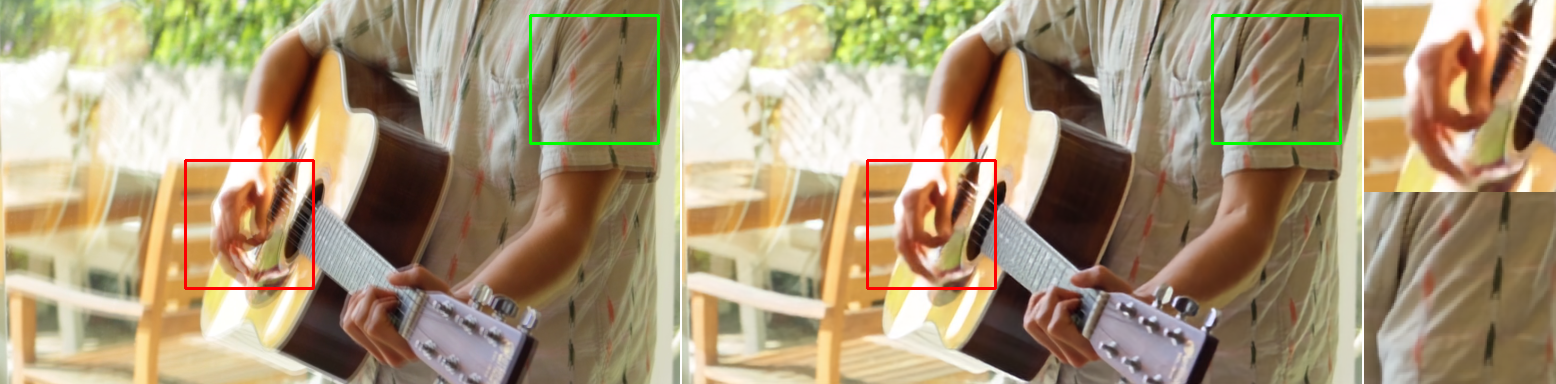}
    \caption{Examples of frame interpolation from our model. The left frame is the a frame overlay of the first and last frame, the middle frame is the interpolated frame, and the right frame is a zoomed-in crop of the red square. From top to bottom, the frames are interpolated at different times, $t=0.5$, $t=0.625$, $t=0.75$, $t=0.5$.}
    \label{fig:examples}
\end{figure*}
\paragraph{Video interpolation} Video interpolation is a well-defined task in deep learning which involves generating intermediate frames between consecutive frames of a low frame-rate video. An obvious application of video interpolation is in creating smooth transitioning slow-motion videos from an original video to fill unknown frames at missing times. Recent state-of-the-art works ~\cite{huang2022real, kong2022ifrnet, zhang2023extracting, jin2023unified} all involve optical flow estimation while warping previous frames to match predicted frames. Indeed we implicitly do something similar which we further elaborate upon in \cref{sec:experiments}.

\paragraph{Mask propagation and tracking}  Mask propagation has also been tackled with ML-based approaches. Similar to the methods used in frame interpolation, attention mechanisms have been used to associate objects in different frames~\cite{yang2021associating} to obtain state-of-the-art performance. Other approaches focus on better encoding of features corresponding to memory in videos~\cite{cheng2021rethinking}, or through using more expressive models to better capture long-term memory in videos~\cite{cheng2022xmem}.

\paragraph{}
Where previous works use neural ODEs to perform video prediction tasks, here we propose using neural CDEs as the natural alternative. Using neural CDEs ensures the continuous nature of the video is built into the model, while additional temporal data can be easily incorporated using splines. We show that using neural CDEs results in an architecture that is simpler and more intuitive than its neural ODE alternatives and other state-of-the-art methods.

\section{Neural CDEs}\label{sec:ncde}

Neural ODEs \cite{chen2018neural} were introduced as a continuous-time analogue of residual neural networks \cite{he2015deep}. Residual layers can be understood as a (discrete-time) difference equation
\begin{equation}
    z_{L+1} - z_L = f_\theta(z_L),
\end{equation}
and \cite{chen2018neural} proposed a natural generalization of this to the continuous-time limit. That is, one aims to learn two functions $g_\theta: \mathbb{R}^{\text{feat}} \to \mathbb{R}^\text{latent}$ and $f_\theta: \mathbb{R} \times \mathbb{R}^{\text{latent}} \to \mathbb{R}^{\text{latent}}$. Together they specify a data generating process
\begin{equation}
Z(0) = g_\theta(x) \qc \dd{Z(t)} = f_\theta(t, Z(t)) \dd{t} \label{eq:diffeq},
\end{equation}
given some input feature vector $x \in \mathbb{R}^{\text{feat}}$.
Similar to residual neural networks, the loss function for neural ODEs is typically defined in terms of the \textit{terminal} value of the ODE $\tilde{y} \equiv h_\theta\qty(Z(T; x))$, where $Z(T; x)$ is the solution of \cref{eq:diffeq} evaluated at a fixed terminal time $T$ for an initial condition $Z(0) = g_\theta(x)$, and $h_\theta$ is some function that maps this final latent vector to a prediction. The loss is then calculated in an identical manner to conventional discrete-time neural networks.

This formalism is a remarkably elegant generalization of residual neural networks. However, we argue that this original formulation does not realize the full power of differential equations. In particular, it makes the somewhat uneconomical decision to marginalize out the path travelled by the latent vector, keeping only its terminal value. This is in some ways natural: the original formulation of neural ODEs as continuous-time residual neural networks led, logically, to the interpretation of intermediate `virtual' times in \cref{eq:diffeq} as merely being intermediate layers of a neural network. In this context, indeed it is unclear how the information encoded in this path can be applied for one-time prediction tasks, such as image classification. However, consider a different context. As hinted at before, for the purposes of modeling video data, one might consider the virtual time axis of the differential equation to coincide with an actual time axis. In this case, the entire path has clear value for almost any task one might care to consider.

The key issue for using neural ODEs for modeling sequential or time series data is that the solution of the ODE (the path $Z(t)$) is fully determined by the initial condition $Z(0)$ leaving no mechanism to update the path based on subsequent data. Working entirely in continuous-time, this problem is solved by neural controlled differential equations \cite{kidger2020neural} which instead take as input a full path $V(t)$ and learn a mapping to a new path $Z(t)$
\begin{equation}\label{eq:cde}
    \dd{Z(t)} = f_\theta(t, Z(t))\dd{V(t)}.
\end{equation}
Neural CDEs can be thought of as the continuous-time recurrent neural networks \cite{kidger2022neural}. We also observe that we can recover a neural ODE by simply setting $V(t) = t$.

Naturally, most real-life data is not stored in continuous time. In the case of videos, a single element of a dataset comprises a collection of frames with associated timestamps (which we do not assume are regularly spaced). Thus the full CDE process for $J$ input frames $x_{1:J}$ depicted in Fig.~\ref{fig:scheme} comprises of four steps:

\begin{enumerate}
    \item Encode into a discrete-time latent space $v_j = \mathcal{E}_\theta(x_j)$ for $j \in \{1,\dots,J\}$.
    \item Use splines \cite{morrill2022choice} to convert into a continuous-time latent path $V(t) = \mathsf{CubicSpline} (v_{1:J})$.
    \item\label{itm:step3} Map to a new continuous-time latent path using the neural CDE, \cref{eq:cde}, $Z(t) = \mathsf{CDE}_\theta(V(t))$.
    \item\label{itm:step4} Discretize and decode $y_{k} = \mathcal{D}_\theta(Z(t_k))$ for $k \in \{1,\dots,K\}$ and associated timestamps $t_{1:K}$.
\end{enumerate}

The exact form of the output $y_{1:K}$ depends on the task at hand, but as it is mapped from a continuous-time latent space, it need not be in the same form as the input $x_{1:J}$ in terms of either data format (e.g., number of frames, frame-rate, or resolution), or stylistic and semantic format.

The cubic splines map the discrete-time latent frames $v_{1:J}$ into a continuous-time latent path. In theory, this step could be replaced with any interpolation method (e.g. polynomial regression or kernel smoothing) or a learnable function. We choose splines, following the literature \cite{kidger2020neural}, as they are computationally efficient and are optimal under a generic smoothness regularisation condition \cite{green1993nonparametric}, in particular we use cubic Hermite splines following \cite{morrill2022choice}. Naturally, as the time-step between input frames decreases the choice of interpolation scheme becomes less important.

The full model (encoder, decoder and CDEs) is then trained end-to-end. The inclusion of the neural CDEs allows the model to learn continuous-time features which are not captured by the splines such as optical flow \cite{barron1994performance} or contextual features, therefore significantly increasing the expressibility of the model.

\begin{figure*}[!h]
    \centering
    \includegraphics[width=\textwidth]{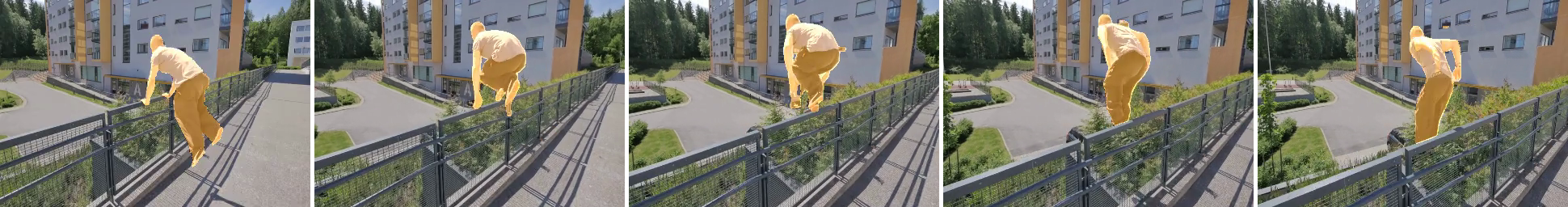}
    \includegraphics[width=\textwidth]{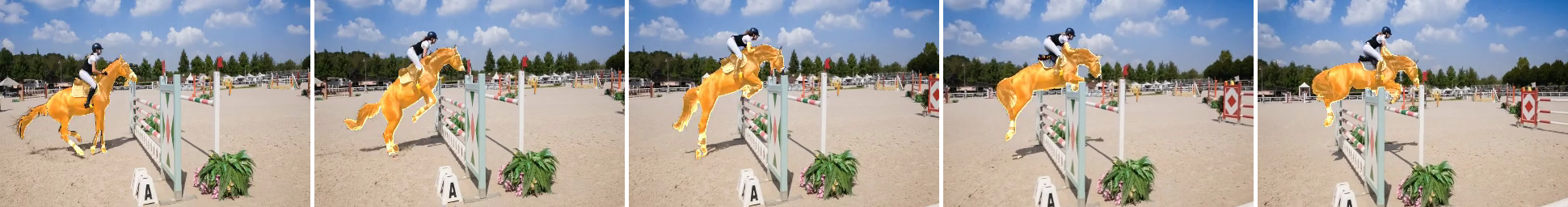}
    \includegraphics[width=\textwidth]{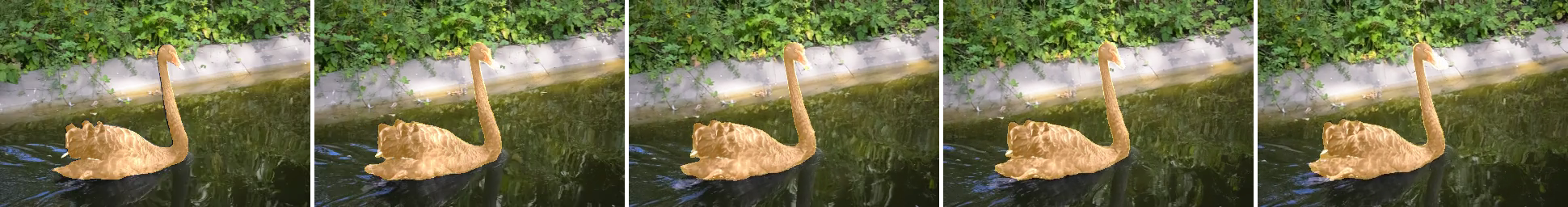}
    \includegraphics[width=\textwidth]{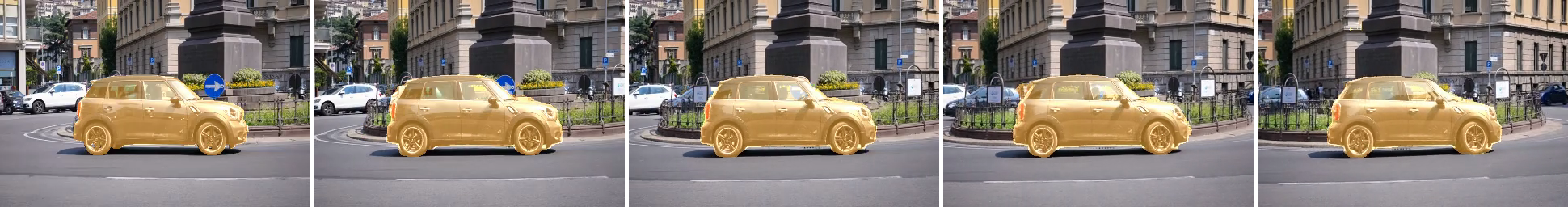}
    \includegraphics[width=\textwidth]{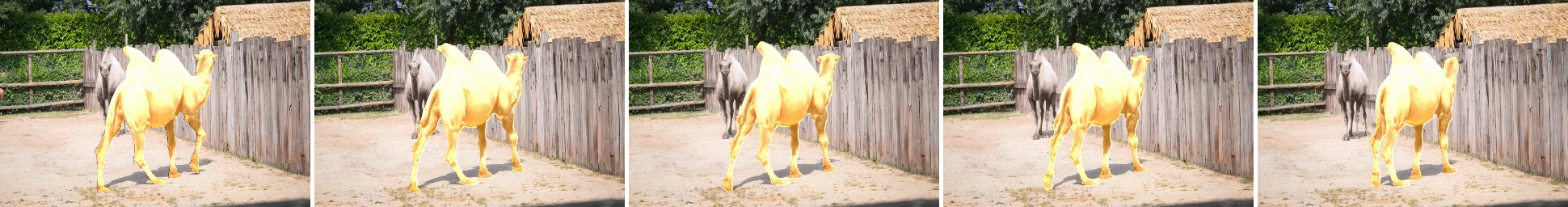}
    \caption{Examples of mask propagation. The first frame on the left represents the ground truth input mask. We show the 4th, 7th, 10th, 13th frames afterwards to display how our method works across longer videos and not just short successive sequences.}
    \label{fig:prop-examples}
\end{figure*}

\begin{figure*}
    \centering
    \includegraphics[width=\textwidth]{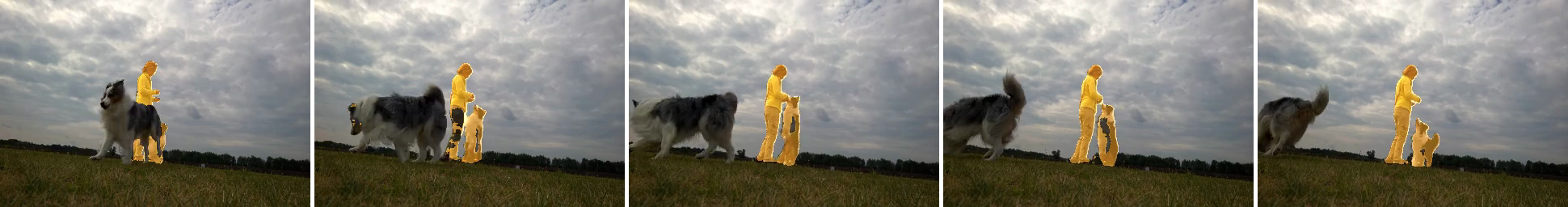}
    \includegraphics[width=\textwidth]{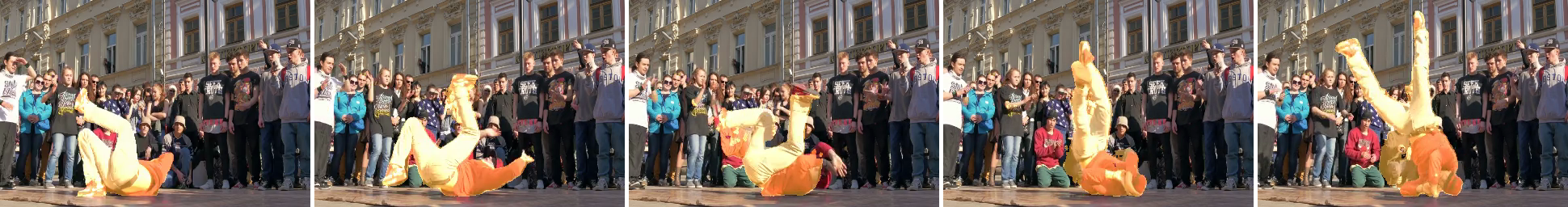}
    \caption{When parts of the initial mask is occluded, our method fails to understand the entire object in some frames (top row, second image). As with all methods, our method also runs into general errors with over and under masking boundaries of objects (bottom row, some in \cref{fig:prop-examples}).}
    \label{fig:prop-occlusions}
\end{figure*}

\section{Experiments}
\label{sec:experiments}

We apply CDEs to two different video tasks in computer vision. Unlike other state-of-the-art methods, our method does not require us to directly learn optical flow. This is implicitly learned in the CDE portion. In turn, this results in a simpler model architecture (see \cref{fig:scheme} for more details), which appears similar to a U-Net, but with all internal components in continuous time.

\paragraph{Video Interpolation} Neural differential equations are powerful due to their ability to model a continuous latent space, therefore, we are able to sample video frames at arbitrary time points. The most obvious extension of this principle is frame interpolation.

A first natural idea would be to directly use splines without the CDE (i.e., decoding $V(t)$ directly in Step \ref{itm:step4}). However, as shown in \cref{table:inteprolation-eval-table}, we find that removing Step \ref{itm:step3} leads to a notable decrease in performance. This suggests learning continuous-time dynamics via CDEs significantly increases the performance of the model.

We vary interpolation factors $n_f\in \{2, 4, 8, 16\}$. An interpolation factor of 2 means one frame generated between every pair of original frames, 4 being three generated frames between every pair, and so on. The number of output frames is defined by $K = n_f \times (J - 1) + 1$ for $J$ input frames. We note that the continuous-time representation of our model permits interpolation at arbitrary timestamps.

Here, we train our model to optimize an $L_1$ loss. A small detail to be noted is that we separate the frames of the video $y$ into two sets of frames $y^{\mathrm{rec}}$ and $y^{\mathrm{int}}$, corresponding to frames that will be reconstructed (the times at which we have known frames) and frames that correspond to interpolated frames (the unknown frames we want to generate), respectively. We weight the $L_1$ loss on $y^{\mathrm{int}}$ heavier since those are the frames to be generated.

\paragraph{Mask Propagation and Tracking} Mask propagation and object tracking throughout a video require the same techniques presented in a frame interpolation task. In particular, by deriving the dynamics of a video, we are able to properly define how objects in subsequent images flow from one frame to the next. Our technique is able to capture the overall flow of objects in a video with a simpler architecture.

We train our model to optimize the binary cross entropy loss.

\subsection{Training}

\paragraph{Video Interpolation} We train our model on Vimeo-90K~\cite{xue2019video} triplets and septuplets. During training, we only predict the middle frames at t = 0.5. During inference, we can predict either sequentially or using a multi-scale recursive pyramid framework where we always predict the middle frame between every pair of frames first. We found that the suggested multi-scale recursive framework works best.

We train our model with a learning rate of 3e-5 and no learning rate scheduler. We use a batch size of 40 for the Vimeo-triplets and 12 for Vimeo-septuplets. Finally, we train our model for $100,000$ steps. 

\begin{table*}[!h]
\centering
\captionof*{table}{\textbf{Interpolation Evaluation on Vimeo90k, UCF-101}}
\begin{tabular}{ ccccc } 
\hline
 \textbf{Method} & \# \textbf{Parameters} & \textbf{Vimeo-Triplets} & \textbf{Vimeo-Septuplets} & \textbf{UCF-101} \\
\textbf{Name} & (Millions) & $\text{PSNR}_{\uparrow}$ $\text{SSIM}_{\uparrow}$ & $\text{PSNR}_{\uparrow}$ $\text{SSIM}_{\uparrow}$ & $\text{PSNR}_{\uparrow}$ $\text{SSIM}_{\uparrow}$ \\
\hline
 RIFE~\cite{huang2022real} & 10.7 & 35.64\ \ \ 0.966 & 30.11\ \ \ 0.904 & 21.46\ \ \ 0.742 \\
 EMA~\cite{zhang2023extracting} & 65.7 & \textbf{36.21}\ \ \ 0.968 & \textbf{32.10}\ \ \ \textbf{0.931} & 22.08\ \ \ \textbf{0.759} \\
 IFRNet~\cite{kong2022ifrnet} & 4.96 & 35.79\ \ \ 0.968  & 25.58\ \ \ 0.827 & 20.51\ \ \ 0.697 \\
 UPRNet~\cite{jin2023unified} & 1.70 & 36.02\ \ \ \textbf{0.969} & 30.08\ \ \ 0.905 & 21.27\ \ \ \textbf{0.759} \\
\hline
Ours w/o CDEs & 22.7 & 29.76\ \ \ 0.863 & 26.93\ \ \ 0.832 & 21.20\ \ \ 0.709 \\
Ours & 43.9 & 35.50\ \ \ 0.951 & \textbf{32.10}\ \ \ 0.925 & \textbf{22.13}\ \ \ 0.746 \\
\hline
\end{tabular}
\caption{Vimeo-Triplets are evaluated on 2x interpolation, Vimeo-Septuplets are evaluated on 4x interpolation, and UCF-101 is evaluated on 8x interpolation.}
\label{table:inteprolation-eval-table}
\end{table*}

\begin{table*}[!h]
\centering
\captionof*{table}{\textbf{Interpolation Evaluation on KTH (Vid-ODE comparison)}}
\begin{tabular}{ cccccc } 
\hline
 \textbf{Method} & \# \textbf{Parameters} & \textbf{KTH ($n_f=2$)} & \textbf{KTH ($n_f=4$)} & \textbf{KTH ($n_f=8$)} \\
\textbf{Name} & (Millions) &  $\text{PSNR}_{\uparrow}$\ \ \ $\text{SSIM}_{\uparrow}$ & $\text{PSNR}_{\uparrow}$\ \ \ $\text{SSIM}_{\uparrow}$ & $\text{PSNR}_{\uparrow}$\ \ \ $\text{SSIM}_{\uparrow}$ \\
\hline
 Vid-ODE~\cite{park2021vid} & 1.91 & 30.05\ \ \ 0.863 & - & - \\
\hline
Ours & 43.9 & \textbf{37.77}\ \ \ \textbf{0.963} & \textbf{35.07}\ \ \ \textbf{0.942} & \textbf{31.07}\ \ \ \textbf{0.865} \\
\hline
\end{tabular}
\caption{We compare our model to Vid-ODE using 2x interpolation factor on the KTH dataset. We trained Vid-ODE model for 500 epochs on the KTH dataset using the code provided by their official repository. The KTH dataset is completely unseen by our model during training.}
\label{table:inteprolation-vid-ode-eval-table}
\end{table*}

\begin{table*}[!h]
\centering
\captionof*{table}{\textbf{Mask Propagation Evaluation on DAVIS 2017}}
\begin{tabular}{ ccccc } 
\hline
 \textbf{Method} & \# \textbf{Parameters} & & \textbf{DAVIS 2017} \\
\textbf{Name} & (Millions) & $\mathcal{J}$ & $\mathcal{F}$ & $\mathcal{J} \& \mathcal{F}$ \\
\hline
 AOT~\cite{yang2021associating} & 14.9 & \textbf{0.828} &0.875 & \textbf{0.851}\\
 STCN~\cite{cheng2021rethinking} & 54.4 & 0.814 & \textbf{0.889} & \textbf{0.851} \\
 XMem~\cite{cheng2022xmem} & 62.2 & 0.807 & 0.884 & 0.845 \\
\hline
Ours & 40.0 & 0.807 & 0.884 &  0.845\\
\hline
\end{tabular}
\caption{Evaluation of our method on DAVIS 2017 dataset.}
\label{table:propagation-eval-table}
\end{table*}

\paragraph{Mask Propagation and Tracking} We train on the YouTube VOS ~\cite{xu2018youtube} and DAVIS 2017~\cite{pont2017davis} datasets. Rather than extracting the labels from the objects, we randomly select a number of masks to track and propagate throughout the video.

Similar to the video interpolation task, we start with a learning rate of 3e-5 and no learning rate scheduler. We train our model for 50,000 steps using a batch size of 6 for DAVIS and YouTube VOS.

\subsection{Results}

\paragraph{Video Interpolation} The neural CDE method performs well on the frame interpolation task as shown in \cref{table:inteprolation-eval-table}. The neural CDE approach significantly outperforms Vid-ODE, which is the method most similar to ours, as shown in \cref{table:inteprolation-vid-ode-eval-table}. We do not perform interpolation with Vid-ODE on KTH using $n_f=4$ or $n_f=8$ as the method requires composition masks and image differences \cite{park2021vid} which are not readily accessible in the provided code for Vid-ODE, but we expect the same conclusion.

\paragraph{Mask Propagation and Tracking} Our method gets near similar results to current state-of-the-art for this task. \footnote{We do not perform evaluation on YouTube VOS given how we do not output labels for objects but only track and propagate the initial masks.} Like other methods, our method also has its own failure cases (as shown in \cref{fig:prop-occlusions}). One shortcoming occurs when we have initial occlusions. This particular issue is often tackled by using temporal attention which relates previous and future frames to one another ~\cite{yang2021associating, cheng2022xmem}. A second shortcoming relates to imprecise generated masks on sharp boundaries. In many cases, this problem is mitigated by using an additional refinement model \cite{zhang2021refinemask}. Investigating the aforementioned solutions in our context could mitigate our shortcomings and represents compelling future research.

\section{Conclusions and Outlook}

In this work, we have explored the potential of controlled differential equations as a tool for video modeling. Our method's architectural simplicity is one of its standout features. This simplicity, however, does not come at the cost of performance. On the contrary, our approach has consistently rivaled, and in certain scenarios, surpassed state-of-the-art techniques on video interpolation and mask propagation tasks. The ability of our method to implicitly learn optical flow is a significant advancement, eliminating the need for explicit optical flow learning which is often complex.

One of the most exciting prospects of our approach is its compatibility with pre-trained image models (encoders and decoders). By fine-tuning these SOTA models for our frame encoders, we can bridge the gap between image and video tasks. This synergy could lead to the development of models that are both efficient to train and accurate for video modeling.

However, like all novel approaches, our method is not without its challenges. Videos with rapid motion and initial occlusions have posed difficulties. We view these not just as challenges but as opportunities for refinement: the foundational principles of our method, especially the smoothness enforced by latent differential equations and modelling as much as possible in continuous-time, provides a robust and natural framework with which these challenges can be addressed via future research. 

As mentioned in \cref{sec:ncde}, it would be desirable to replace the splines with a learnable function. With the recent success of attention and transformers, an interesting avenue for future research would be to leverage these ideas within our continuous-time framework.

Potential future applications for our method are numerous. For instance, we can use CDEs as a foundation for video-to-models. A natural application of this is real-time video stylization, where the underlying CDE will implicitly enforce continuity across frames in the stylized video. In addition, the principles of CDEs can be applied to develop more efficient video compression techniques, ensuring high-quality videos at significantly reduced sizes.

\bibliographystyle{unsrt}
\bibliography{references}

\end{document}